\def\year{2020}\relax
\documentclass[letterpaper]{article}

\usepackage{times}
\usepackage{epsfig}
\usepackage{graphicx}
\usepackage{amsmath}
\usepackage{amssymb}
\usepackage{color}
\usepackage{microtype}
\usepackage{algorithm}
\usepackage{algorithmic}
\usepackage{ifthen}
\usepackage{booktabs}
\usepackage{multirow}
\usepackage{enumitem}
\usepackage{booktabs}
\usepackage[switch]{lineno}
\usepackage{helvet}
\usepackage{courier}
\usepackage[hyphens]{url}
\usepackage{aaai20}

\newcommand{\commentout}[1]{} 

\newcommand{\patch}{p}
\newcommand{\transformation}{t}
\newcommand{\params}{\theta} 
\newcommand{\transformations}{T}
\newcommand{\model}{\mathbf{W}}
\newcommand{\loss}{\mathcal{L}}
\newcommand{\diff}{\mathcal{D}}
\newcommand{\degradationloss}{\loss_{\mathrm{deg}}}
\newcommand{\tripletloss}{\loss_{\mathrm{trp}}}
\newcommand{\totalloss}{\loss}
\newcommand{\minibatch}{B}

\usepackage{graphicx}
\title{Revisiting Image Aesthetic Assessment via Self-Supervised Feature Learning}
\author{\textbf{Kekai Sheng\textsuperscript{\rm 1,2}, Weiming Dong\textsuperscript{\rm 1}\thanks{Corresponding author}, Menglei Chai\textsuperscript{\rm 3}, Guohui Wang\textsuperscript{\rm 3}, Peng Zhou\textsuperscript{\rm 4},} \\
\Large \textbf{Feiyue Huang\textsuperscript{\rm 2}, Bao-Gang Hu\textsuperscript{\rm 1}, Rongrong Ji\textsuperscript{\rm 5}, Chongyang Ma\textsuperscript{\rm 6}} \\
\textsuperscript{\rm 1} Institute of Automation, Chinese Academy of Sciences,
\hspace{5pt}
\textsuperscript{\rm 2} Youtu Lab, Tencent,
\hspace{5pt}
\textsuperscript{\rm 3} Snap Inc. \\
\textsuperscript{\rm 4} North China Electric Power University,
\hspace{5pt}
\textsuperscript{\rm 5} Xiamen University, China,
\hspace{5pt}
\textsuperscript{\rm 6} Kuaishou Technology \\
\{saulsheng, garyhuang\}@tencent.com, \{weiming.dong, baogang.hu\}@ia.ac.cn, \{cmlatsim, robertwgh\}@gmail.com \\
pzhou@ncepu.edu.cn, rrji@xmu.edu.cn, chongyangma@kwai.com
}

\begin{document}

\maketitle

\begin{abstract}
Visual aesthetic assessment has been an active research field for decades.
Although latest methods have achieved promising performance on benchmark datasets, they typically rely on a large number of manual annotations including both aesthetic labels and related image attributes.
In this paper, we revisit the problem of image aesthetic assessment from the self-supervised feature learning perspective.
Our motivation is that a suitable feature representation for image aesthetic assessment should be able to distinguish different expert-designed image manipulations, which have close relationships with negative aesthetic effects.
To this end, we design two novel pretext tasks to identify the types and parameters of editing operations applied to synthetic instances.
The features from our pretext tasks are then adapted for a one-layer linear classifier to evaluate the performance in terms of binary aesthetic classification.
We conduct extensive quantitative experiments on three benchmark datasets and demonstrate that our approach can faithfully extract aesthetics-aware features and outperform alternative pretext schemes.
Moreover, we achieve comparable results to state-of-the-art supervised methods that use 10 million labels from ImageNet.
\end{abstract}

\section{Introduction}

With the explosive growth of online visual data, the demand for image aesthetic assessment~\cite{datta2006studying} in many multimedia applications has been dramatically increased. Typically, this assessment process seeks to evaluate the aesthetic level of each image according to certain rules commonly agreed by human visual perception, ranging from fine-grained local textures and lighting details to high-level semantic layout and composition. These highly subjective and ambiguous perceptual metrics pose formidable challenges on designing intelligent agents to automatically and quantitatively measure image aesthetics, especially with conventional hand-crafted features.

Following the recent advances in deep convolutional neural networks, researchers have explored various data-driven learning based approaches for aesthetic assessment and have reported impressive results in the past few years~\cite{ma2017lamp,mai2016composition,sheng2018attention,talebi2018nima}, benefiting from several image aesthetic benchmarks~\cite{murray2012ava,kong2016photo}. However, the inherent shortcomings of these datasets are still deterring us from continuously scaling up the volume of training data and improving the performance: (1) the sizes of existing image aesthetic datasets are far from enough to feed up latest neural networks with very deep architectures, since the labor work involved in manual labeling is prohibitively expensive and developing aesthetics-invariant data augmentation solution remains an open problem; (2) meanwhile, subjective human annotations are often strongly biased towards personal aesthetic preference, and thus require excessive amount of data to neutralize such inconsistency for reliable training.

\begin{figure}
    \centering
    \includegraphics[width=\linewidth]{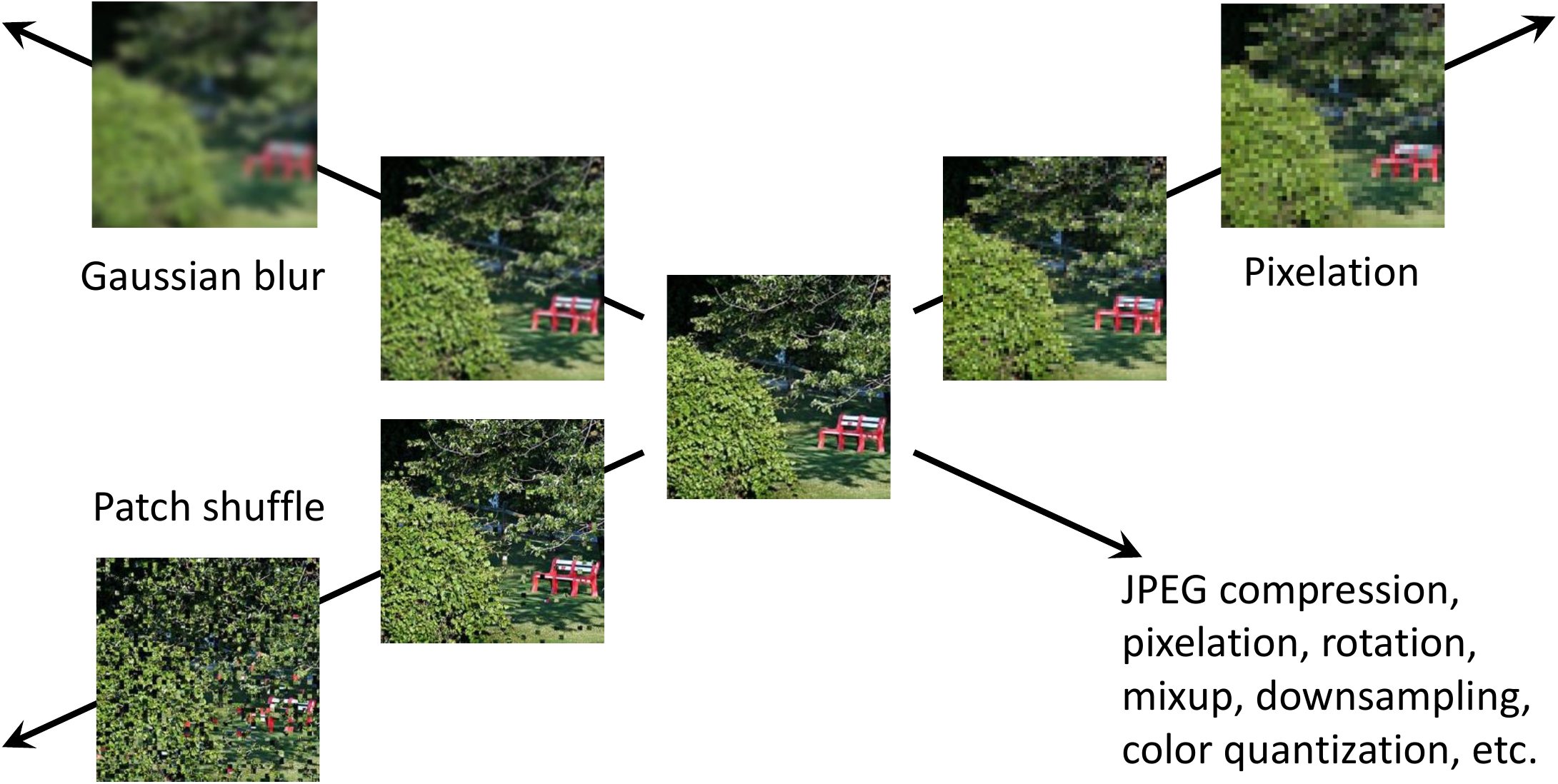}
    \caption{Illustration of our key idea. Some image editing operations typically have negative aesthetic impact and the degree of the impact is related to manipulation parameters.}
    \label{fig:Operations_toward_Aesthetics}
\end{figure}

Due to the poor scalability and consistency of available aesthetics datasets, it is becoming more and more attractive to break these bottlenecks of purely supervised training with unsupervised~\cite{hinton2006reducing,vincent2008extracting}, or more specifically, self-supervised features~\cite{de1994learning,doersch2015unsupervised,pathak2016context,zhang2016colorful}. The main idea of self-supervision is to design pretext tasks that are naturally available, contextually relevant, and capable of providing proxy loss as the training signal to guide the learning of features which are consistent with the real target. As a result, we can significantly broaden the scope of training data without introducing any further annotation cost.

In the context of image aesthetic assessment, due to the absence of an in-depth understanding of the human perception system, designing such a proper self-supervised pretext task can be challenging. However, although it is difficult to quantitatively measure the aesthetic score, predicting the relative influence of certain controlled image editing operations can be much easier.
Therefore, we propose two key observations to help design the task (see Fig.~\ref{fig:Operations_toward_Aesthetics}): (1) some parametric image manipulation operations, such as blurring, pixelation, and patch shuffling, are very likely to have consistent negative impact on image aesthetics; (2) and the degree of impact due to such degradation will be monotonically increasing with respect to values of the corresponding operation parameters.

In this work, motivated by the strong correlation between image aesthetic levels and some degradation operations, we propose, to the best of our knowledge, the very first self-supervised learning scheme for image aesthetic assessment.
The core idea behind our approach is to extract aesthetics-aware features with two novel self-supervision pretext tasks on distinguishing the type and strength of image degradation operations.
To improve the training efficiency, we also introduce an entropy-based weighting strategy to filter out image patches with less useful training signals.
The experimental results demonstrate that our self-supervised aesthetics-aware feature learning is able to achieve promising performance on available aesthetics benchmarks such as AVA~\cite{murray2012ava} and AADB~\cite{kong2016photo}, and outperforms a wide range of commonly adopted self-supervision schemes, including context prediction~\cite{doersch2015unsupervised}, jigsaw puzzle~\cite{noroozi2016unsupervised}, colorization~\cite{larsson2017colorization,zhang2016colorful}, and rotation recognition~\cite{gidaris2018unsupervised}.

In summary, our main contributions include:
\begin{itemize}
\setlength{\itemsep}{0pt}
\setlength{\parskip}{0pt}
\setlength{\parsep}{0pt}
\item We propose a simple yet effective self-supervised learning scheme to extract useful features for image aesthetic assessment without using manual annotations.

\item We present an entropy-based weighting strategy to help strengthen meaningful training signals in manipulated image patches\commentout{ and promote training efficiency}.

\item On three image aesthetic assessment benchmarks, our approach outperforms other self-supervised counterparts and even works better than models pre-trained on ImageNet or Places datasets using a large number of labels.
\end{itemize}

\section{Related Work}

\paragraph{Image aesthetic assessment} has been extensively studied in the past two decades.
Conventional solutions use hand-crafted feature extractors designed with domain expertise to model the aesthetic aspects of images~\cite{datta2006studying,ke2006the}. More recently, the power of deep neural networks makes it possible to learn feature representations that can surpass hand-crafted ones.
Typical approaches include distribution-based objective functions~\cite{talebi2018nima,jin2018predicting}, multi-level spatially pooling~\cite{hosu2019effective}, attention-based learning schemes~\cite{sheng2018attention}, and attribute/semantics-aware models~\cite{ma2017lamp,pan2019image}.
The advance of learning-based image aesthetic assessment has also inspired a number of practical solutions for various usage scenarios such as clothing~\cite{yu2018aesthetic} and food~\cite{Sheng:2018:GPD}.

\paragraph{Self-supervised feature learning} can be considered as one type of unsupervised learning algorithms~\cite{de1994learning}, which intends to learn useful representations without manual annotations.
Many effective pretext tasks have been proposed in this direction, such as context prediction~\cite{doersch2015unsupervised}, colorization~\cite{zhang2016colorful}, split-brain~\cite{zhang2017split}, and RotNet~\cite{gidaris2018unsupervised}.
The representations learned from self-supervised schemes turn out to be useful for many downstream tasks, \emph{e.g.}, tracking~\cite{vondrick2018tracking}, re-identification~\cite{fan2018unsupervised}, and image generation~\cite{lucic2019high-fidelity}\commentout{, especially in the case of using fewer labels}.

\paragraph{Aesthetics-aware image manipulations} can be generally divided into two categories:
(1) Rule-based approaches leverage empirical knowledge and domain expertise to enforce well-established photographic heuristics such as gamma correction and histogram equalization.
(2) Data-driven approaches focus on learning powerful feature representations from examples to improve aesthetics in certain aspects, such as aesthetics-aware image enhancement~\cite{deng2018aesthetic,hu2018exposure}, image blending~\cite{Hung_blend_2018}, and colorization~\cite{zhang2016colorful}.

\section{Our Method}
\label{sec:method}

\subsection{Key Observations}
\begin{figure*}
    \centering
    \begin{minipage}[c]{.7\textwidth}
        \centering
        \includegraphics[width=0.85\linewidth]{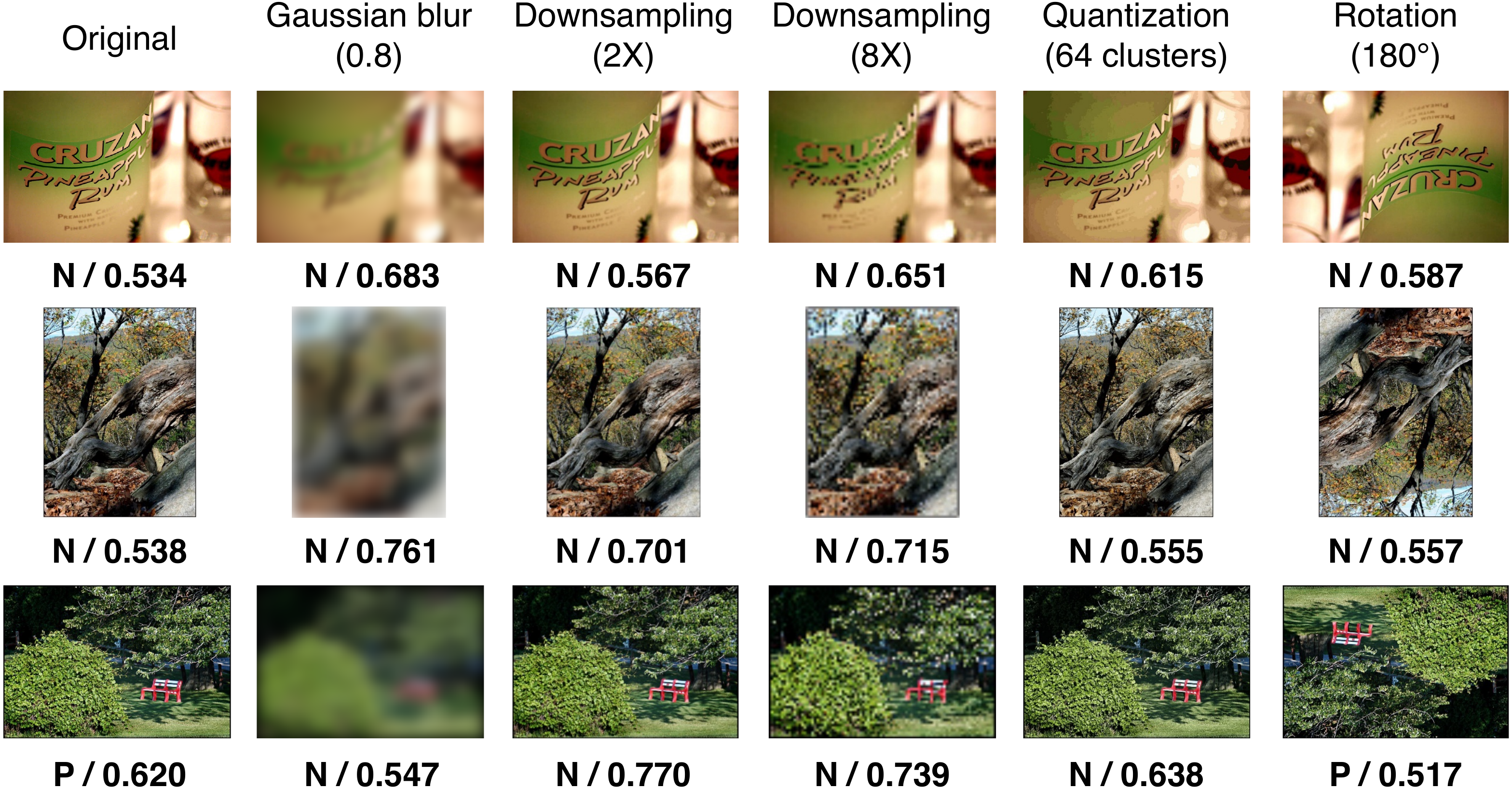}
    \end{minipage}
    \hfill
    \begin{minipage}[c]{.25\textwidth}
        \centering
        \caption{Some learning-free parametric image editing operations can introduce controllable aesthetic degradations on input images.
        The aesthetic label (\textbf{P} for positive and \textbf{N} for negative), as well as the corresponding assessment confidence predicted by a fully supervised model are listed beneath each image.}
        \label{fig:FineGrainedAesthetic}
    \end{minipage}
\end{figure*}

Our self-supervised feature learning approach for image aesthetic assessment is based on two key observations.
First, experiments in previous work on image aesthetic assessment benchmarks indicate that inappropriate data augmentation (e.g., brightness/contrast/saturation adjustment, PCA jittering) during the training process will result in performance degradation at the test time~\cite{murray2012ava,kong2016photo}.
Second, it is observed that a convolutional neural network (CNN) model trained from manually annotated aesthetic labels can inherently acquire the ability to distinguish fine-grained aesthetic differences caused by various image manipulation methods.
In Fig.~\ref{fig:FineGrainedAesthetic}, for instance, some image editing operations (such as Gaussian blur, downsampling, color quantization, and rotation) can increase the prediction confidence of aesthetically negative images, or even turn an aesthetically positive example into a negative one (as shown in the last row of Fig.~\ref{fig:FineGrainedAesthetic}).

Consequently, we argue that these fine-grained perceptual quality problems, without manual annotations, are closely related to the image aesthetic assessment task, for which meaningful training instances can be constructed via proper expert-designed image manipulations.

\subsection{Selected Image Manipulations}
\label{sec:selected_image_manipulations}

According to empirical knowledge, different image editing operations have diverse effects on the manipulated output~\cite{ma2018end2end,zhang2018perceptual}.
Furthermore, some operations require complex parameter settings (e.g., inpainting) and it is often difficult to automatically compare the output to the input in terms of aesthetic level (e.g., grayscale conversion).
In our case, however, we need to select image manipulations with easily controllable parameters and predictable perceptual quality.

Specifically, as listed in Table~\ref{Tab:DegradationType}, we adopt a variety of image manipulation operations with different parameters for artificial training instances, including (1) downsampling by a scaling factor and upsampling back to the original resolution via bilinear interpolation; (2) JPEG compression with a percentage number to control the quality level; (3) Gaussian noise controlled by the variance; (4) Gaussian blur controlled by the standard deviation; (5) color quantization into a small number of levels; (6) brightness change based on a scaling factor; (7) random patch shuffle~\cite{kang2017patchshuffle}; (8) pixelation based on a patch size; (9) rotation by a certain degree; and (10) linear blending (mixup~\cite{zhang2017mixup}) based on a constant alpha value.

\begin{table}
    \centering
    \caption{Image editing operations and the parameters adopted to construct meaningful pretext tasks.
    }
    \setlength{\tabcolsep}{1mm}{
    \footnotesize{
    \begin{tabular}{c|lc}
    \toprule
    Attribute & Operation & Parameters \\
    \hline
    \hline
    \multirow{2}{*}{Much noise} & JPEG compression & $\{60, \, 10\}$ \\
     & Gaussian noise  & $\{0.2, \, 0.8 \}$\\
    \hline
    Camera Shake & Rotation & $\{90, \, 180, \, 270\}$ \\
    \hline
    \multirow{3}{*}{Soft / Grainy} & Downsampling & $\{4, \, 6\}$\\
     & Quantization & $\{64, \, 8\}$ \\
     & Pixelation & $\{4, \, 8\}$ \\
    \hline
    Poor lighting & Exposure & $\{0.5, \, 3.0\}$ \\
    \hline
    Fuzzy & Gaussian blur  & $\{0.2, \, 0.8 \}$\\
    \hline
    \multirow{2}{*}{Distracting} & Patch shuffle & $\{0.1, \, 0.5 \}$ \\
     & Mixup & $\{0.1, \, 0.4\}$ \\
    \bottomrule
    \end{tabular}
    }
    }
    \label{Tab:DegradationType}
\end{table}

\subsection{Aesthetics-Aware Pretext Tasks}
In this section, we propose our aesthetics-aware pretext task in a self-supervised learning scheme from two aspects.
On one hand, the ability to categorize different types of image manipulations can be beneficial to the learning of aesthetics-aware representation.
On the other hand, for the same type of editing operation, different control parameters can render various aesthetic levels correspondingly, and the tendency of quality shift is predictable~\cite{marchesotti2015discovering}.
Take JPEG compression for instance, decreasing the output image quality parameter always decreases the aesthetic level.
Therefore, by constructing images with manipulations resulting in predictable degradation behaviors, we can extract meaningful training signals for the fine-grained aesthetics-aware pretext tasks.

\paragraph{Degradation identification loss.}
We denote an image patch as $\patch$, and the manipulation parameter as $\params_{\transformation}$ for $m(\cdot, \params_{\transformation})$.
The loss term of our first pretext task, \emph{i.e.}, $\degradationloss(\patch, \transformation)$, reinforces the model to recognize which operation $t$ has been applied to $\patch$:
\begin{equation}
    \centering
    \begin{split}
        \degradationloss(\patch, \params_{\transformation}) &= - \log P_{\transformation}(\patch; \model), \\
        P_{\transformation}(\patch; \model) &= P(\tilde{\transformation} = \transformation \, | \, m(\patch, \params_{\transformation}); \model),
    \end{split}
    \label{equ:L_deg}
\end{equation}
where $m(\patch, \params_{\transformation})$ is the transformed output patch given the image patch $\patch$ by the parameters $\params_{\transformation}$, and $P(\tilde{\transformation} = \transformation \, | \, m(\patch, \params_{\transformation}); \model)$ is the probability predicted by our model $\model$ that $p$ has undergone a degradation operation of type $\tilde{\transformation}$.

For a comprehensive coverage of image attributes (e.g., resolution, color, spatial structures), we leverage a variety of typical manipulation operations $\transformations$ as listed in Table~\ref{Tab:DegradationType}.
To take better advantage of synthesized instances, we adopt parameters that will always induce aesthetic degradation of observably different patterns.
Apart from these distortions, \textit{None} operation is also taken into consideration.
That is, we require the model to categorize $22$ different classes of editing operations.

\begin{figure*}
\centering
    \begin{minipage}[c]{.67\textwidth}
        \centering
        \includegraphics[width=0.92\linewidth]{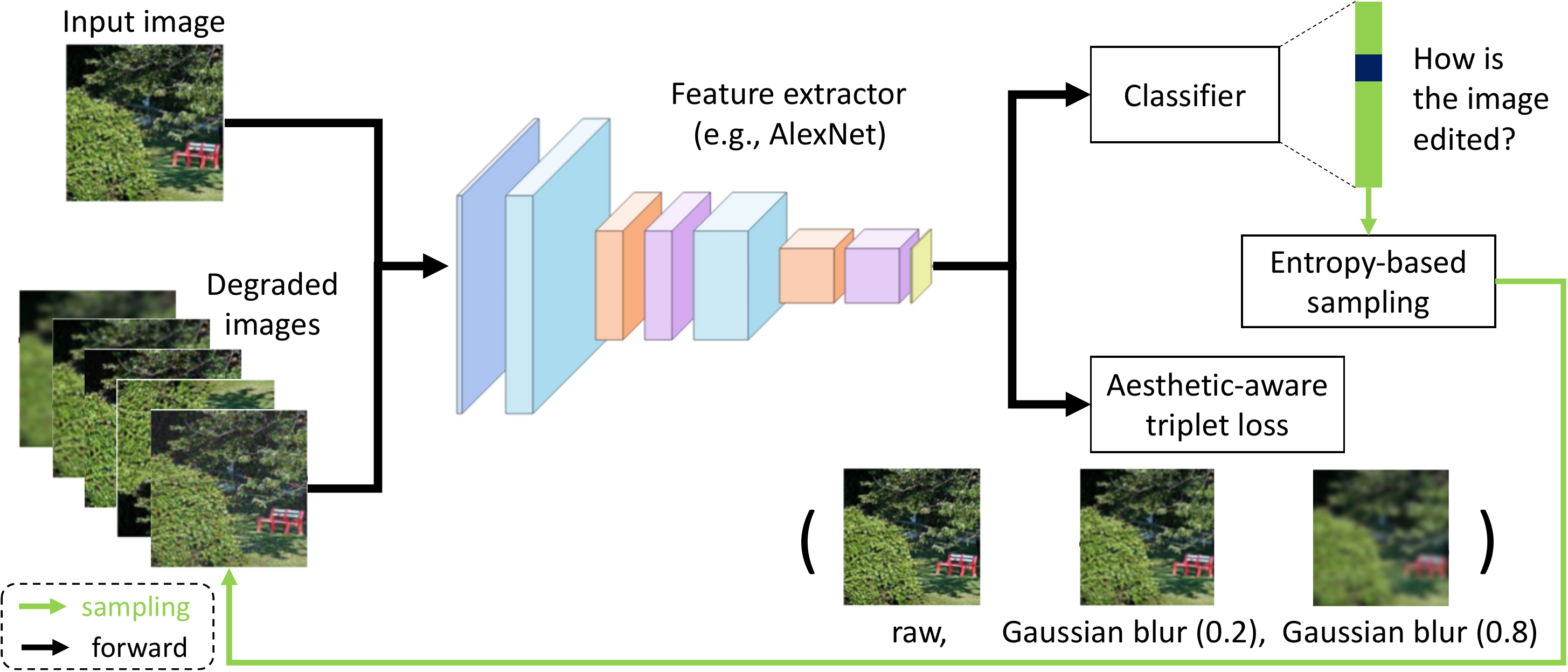}
    \end{minipage}
    \hfill
    \begin{minipage}[c]{.30\textwidth}
        \centering
        \caption{The diagram of our self-supervised approach.
        We propose two aesthetic-aware pretext tasks, and apply an entropy-based weighting scheme to enhance the training efficiency.
        In this way, we learn useful aesthetic-aware features in a label-free manner.}
        \label{fig:SSAVAPretextTraining}
    \end{minipage}
\end{figure*}

\paragraph{Triplet loss.}
Training with $\degradationloss(\patch, \transformation)$ alone is not enough for the task of aesthetic assessment, since some editing operations may produce some low-level artifacts that are easy to detect and will fool the network to learn some trivial features of no practical semantics~\cite{gidaris2018unsupervised}.
Our solution to address this issue is to encode the information via triplets $(\patch, m(\patch, \params_{\transformation_{1}}), m(\patch, \params_{\transformation_{2}}))$, where $\transformation_{1}$ and $\transformation_{2}$ are two different parameters of a certain operation in Tab.~\ref{Tab:DegradationType} (except for rotation and exposure).
The two parameters are specified to create aesthetic distortions with a predictable relationship, \emph{i.e.}, the edited image patch $m(\patch, \params_{\transformation_{1}})$ using $\transformation_{1}$ is aesthetically more positive than $m(\patch, \params_{\transformation_{2}})$.
Therefore, in an ideal aesthetic-aware representation, the distance between the original image patch $\patch$ and $m(\patch, \params_{\transformation_{1}})$ should be smaller than the distance between $\patch$ and $m(\patch, \params_{\transformation_{2}})$.
In this way, we propose the second task, $\tripletloss(\patch, \params_{\transformation_{1}}, \params_{\transformation_{2}})$:
\begin{align}
    \tripletloss(\patch, \params_{\transformation_{1}}, \params_{\transformation_{2}}) &= \max \{ 0, \diff(\patch, \params_{\transformation_{1}}) - \diff(\patch, \params_{\transformation_{2}}) + 1 \}, \\ \nonumber
    \diff(\patch, \params_{\transformation}) &= \Big|\Big| h(\patch, \model) - h(m(\patch, \, \params_{\transformation}), \model) \Big|\Big|_{2}^{2},
    \label{equ:D_triplet}
\end{align}
where $h(\patch, \model)$ is the normalized feature extracted from the model $\model$ given a patch $\patch$\commentout{ and $1$ is for the necessary margin}.

It should be noted that we do not apply the triplet loss term from the beginning of the training process, since the representation learned from $\degradationloss(\patch, \transformation)$ in the early stage can oscillate drastically and thus may not be suitable for comparisons of fine-grained features.
To combat the training dynamics, we activate $\tripletloss(\patch, \params_{\transformation_{1}}, \params_{\transformation_{2}})$ after the curve of $\degradationloss(\patch, \transformation)$ has shown some plateau.

\paragraph{Total loss function.}
By putting the degradation identification loss item and the triplet loss item together, we formulate a new self-supervised pretext task as below:
\begin{equation}
    \centering
        \totalloss = \frac{1}{|\minibatch| \cdot |\transformations|} \sum_{\patch \in \minibatch} \sum_{\transformation, \transformation_{1}, \transformation_{2} \in \transformations} \degradationloss(\patch, \params_{t})
          + \lambda \cdot \tripletloss(\patch, \params_{\transformation_{1}}, \params_{\transformation_{2}}),
    \label{equ:SSAVA}
\end{equation}
where $\minibatch$ is a mini-batch made up of $|\minibatch|$ patches, $\transformations$ is the set of all the image manipulations, and $\lambda$ is a scaling factor to balance the two terms.
The diagram of our proposed self-supervised scheme is shown in Fig.~\ref{fig:SSAVAPretextTraining}.

\paragraph{Entropy-based weighting.}
To improve the training efficiency and reduce noisy signals from misleading instances, we present a simple yet effective entropy-based weighting strategy.
Specifically, after warm up for several epochs, we apply an entropy-based weight $w_{p}$ for each patch $p$:
\begin{equation*}
    w_{p} = \max \{1 + \alpha \cdot \sum_{\transformation} P_{\transformation}(\patch; \model) \cdot \log P_{\transformation}(\patch; \model), 0 \},
\end{equation*}
where $\alpha$ is used to control the lower bound of $w_{p}$ and we set it as $5/3$ in our experiments.
The rationale behind this strategy is that instances of higher entropy values tend to have more uncertain visual cues for image aesthetics, and thus should be assigned with lower weights in the optimization process.

\paragraph{Discussions.}
Our proposed pretext task is similar to \cite{dosovitskiy2014discriminative,liu2017rankiqa} in which CNNs are trained to discriminate instances from different types of data augmentation methods.
Different from these previous methods, we select a set of image manipulation operations and design our loss function carefully so that the learned representation can be aware of significant patterns of visual aesthetics.
Besides, instead of building on selected high quality photos, we conduct pretext task optimization directly on images from ImageNet~\cite{deng2009imagenet}.
This strategy makes our learning scheme more flexible to use.

One might argue that we ignore some global aesthetic factors, \emph{e.g.}, rule of thirds.
The reasons are two folds.
First, global image attributes are more complex to manipulate than local ones which are considered in our approach.
Second, global factors generally involve semantics which are not available in the context of self-supervised feature learning.
We are not intended to mimic the statistics of \textit{real-world} visual aesthetics, but to propose pretext tasks which are suitable for visual aesthetic assessment\commentout{ and learn features \emph{using no manual labels}}.

\section{Experiments}

\subsection{Baseline Methods}
\label{sec:experimental_setup}

We compare the performance of our method with five typical self-supervised visual pretext tasks, as listed below:
\begin{description}
    \item[Context:]
    The context predictor~\cite{doersch2015unsupervised} that predicts the relative positions between two square patches cropped from one input image.
    \item[Colorization:]
    The image colorization task~\cite{zhang2016colorful} requires the model to estimate the color channels from a gray-scale image.
    \item[Split-brain:]
    The cross-channel predictor~\cite{zhang2017split} estimates one subset of the color channels from another with constrained spatial regions.
    \item[Counting:]
    The primitive counting task~\cite{noroozi2017representation} requires the model to generate visual primitives with their number invariant to transformations including both scaling and tiling.
    \item[RotNet:]
    A rotation predictor~\cite{gidaris2018unsupervised} is trained to recognize the 2D rotation applied to input images, i.e., $m(\cdot; \ y)$ is image rotation and $y \in \{ 0^{\circ}, 90^{\circ}, 180^{\circ}, 270^{\circ}\}$.
\end{description}
In our experiments, we use the pre-trained models released by the authors for reliable and fair comparisons\footnote{Context: \url{https://github.com/cdoersch/deepcontext}.}\footnote{Colorization: \url{https://github.com/richzhang/colorization}.}\footnote{Split-brain: \url{https://github.com/richzhang/splitbrainauto}.}\footnote{Counting: \url{https://github.com/gitlimlab/Representation-Learning-by-Learning-to-Count}.}\footnote{RotNet: \url{https://github.com/gidariss/FeatureLearningRotNet}.}.

Additionally, we also testify with three typical pre-training strategies, including the method pre-trained with $1,000$-way object labels from ImageNet~\cite{deng2009imagenet} or $365$-way scene labels from Places~\cite{zhou2017places}, and the Gaussian random initialization \emph{i.e.}, without any pretext task.

\subsection{Training Pipeline}
\label{sec:training_algorithm}

\begin{figure}
    \centering
    \includegraphics[width=\linewidth]{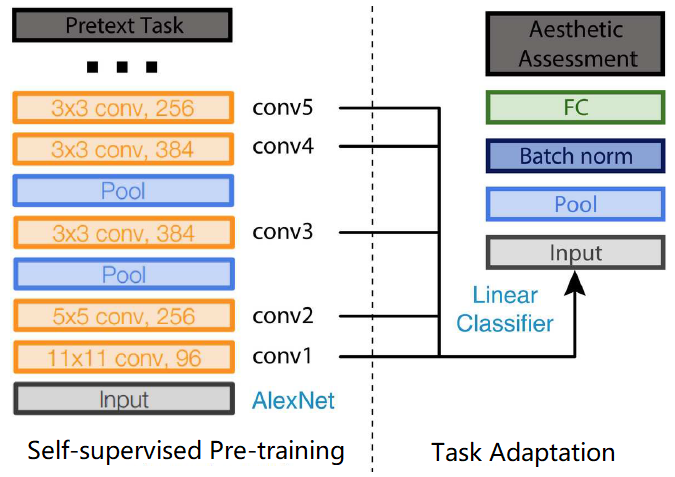}
    \caption{The schematic of a self-supervised learning framework, which we apply in our experiments to evaluate various pretext tasks for image aesthetic assessment.}
    \label{fig:PipelineOfSSAVA}
\end{figure}

Our entire training pipeline (Fig.~\ref{fig:PipelineOfSSAVA}) contains two parts: a {\em self-supervised pre-training} stage to learn visual features with unlabeled images and a {\em task adaptation} stage to evaluate how the learned features perform in the task of image aesthetic assessment.

In the pre-training stage (Fig.~\ref{fig:PipelineOfSSAVA} left), the first few layers share the same structure with AlexNet~\cite{krizhevsky2012imagenet} for fair comparisons.
In the task adaptation stage (Fig.~\ref{fig:PipelineOfSSAVA} right), following the same configurations in~\cite{zhang2017split}, we freeze each model learned in the pre-training and leverage similar linear classifiers
\commentout{ (see Tab.~\ref{Tab:LinearClassifierConfiguration})} to evaluate visual features from each convolutional layer for the task of binary aesthetic classification.
The channel number of each convolutional layer is shown in Fig.~\ref{fig:PipelineOfSSAVA}, and the dimensions of the corresponding fully connected layers are $9600 \times 2$, $9216\times 2$, $9600\times 2$, $9600\times 2$, and $9216\times 2$, respectively.

\paragraph{Implementation details.}
In the pre-training stage, we first resize the shorter edge of each input image to $256$.
Then we randomly crop one patch of resolution $227 \times 227$ from the resized image.
Next, we randomly choose three manipulation operations\commentout{ due to the limited GPU memory} in Tab.~\ref{Tab:DegradationType} to edit each patch.
We apply SGD optimization using a batch size of $64$, with the Nesterov momentum of $0.9$ and the weight decay of $5\mathrm{e}{-4}$.
We begin with a learning rate of $0.1$, dropped it by a factor of $0.2$ after every $10$ epochs.
To eschew training oscillating, we activate $\tripletloss(\patch, \transformation)$ with $\lambda$ of $0.02$ after the first $30$ epochs.
The following adaptation stage shares the same settings except that the learning-rate starts from $0.01$.


\subsection{Benchmarks for Aesthetic Assessment}
\label{sec:benchmarks}

\paragraph{Aesthetic Visual Analysis (AVA).}
The AVA dataset~\cite{murray2012ava} contains approximately $250,000$ images. 
Each image has about $100$ crowdsource aesthetic ratings in the range of $1$ to $10$.
Following the common practice in~\cite{ma2017lamp,hosu2019effective}, we consider images whose average aesthetic scores are no less than $5.0$ as positive instances and adopt the same training/test partition, i.e., $230,000$ images for training and $20,000$ for testing.

\paragraph{Aesthetics with Attributes Database (AADB).}
The AADB dataset~\cite{kong2016photo} contains $10,000$ images with aesthetic ratings $\{ r \in [0, 1] \}$ and eleven additional attributes.
We follow~\cite{kong2016photo} to split the dataset into three partitions, \emph{i.e.}, $8500$, $500$, and $1000$ images for training, validation, and testing, respectively.
Without loss of generality, we binarize the aesthetic ratings into two classes using a threshold of $0.5$, similar to AVA.

\paragraph{Chinese University of Hong Kong-Photo Quality Dataset (CUHK-PQ).}
The CUHK-PQ dataset~\cite{luo2011content} contains $17,690$ images with binary aesthetic labels.
Commonly used training/testing partitions on this dataset include a random $50/50$ split and a five-fold split for cross-validation.
We use the former one in our experiments.

\begin{table*}
    \centering
    \caption{Task generalization performance ($\%$) of different convolutional layers from models guided by different pretext tasks, measured for visual aesthetic assessment with linear layers on the AVA, AADB, and CUHK-PQ benchmarks.
    }
    \scriptsize{
    \setlength{\tabcolsep}{1.1mm}{
    \begin{tabular}{r|cccccc|cccccc|cccccc}
    	\toprule
    	\multirow{2}{*}{\footnotesize{Method}} & \multicolumn{6}{c|}{\footnotesize{AVA}} & \multicolumn{6}{c|}{\footnotesize{AADB}} & \multicolumn{6}{c}{\footnotesize{CUHK-PQ}} \\
    	\cline{2-19}
    	 & conv1 & conv2 & conv3 & conv4 & conv5 & average & conv1 & conv2 & conv3 & conv4 & conv5 & average & conv1 & conv2 & conv3 & conv4 & conv5 & average \\
    	\hline
    	\footnotesize{ImageNet label} & 79.3 & 79.0 & 79.3 & 79.1 & 79.4 & 79.22 & 63.6 & 65.0 & 64.2 & 67.5 & 64.2 & 64.9 & 77.8 & 83.3 & 84.1 & 84.2 & 83.1 & 82.5 \\
    	\footnotesize{Places label} & 79.6 & 79.5 & 79.2 & 79.6 & 79.8 & 79.54 & 61.8 & 63.2 & 63.4 & 65.0 & 65.2 & 63.72 & 76.0 & 82.2 & 82.5 & 83.1 & 81.6 & 81.08 \\
    	\hline
    	\footnotesize{Without pretext} & 77.2 & 77.8 & 78.0 & 78.2 & 78.0 & 77.84 & 61.3 & 58.9 & 62.1 & 61.6 & 64.0 & 61.58 & 73.6 & 72.0 & 73.3 & 74.6 & 73.1 & 73.32 \\
    	\hline
    	\footnotesize{Context} & \underline{79.8} & 79.2 & 79.2 & 79.0 & 79.0 & 79.24 & 59.4 & 59.8 & 62.4 & 62.6 & 63.2 & 61.48 & 69.2 & 75.9 & 77.9 & 79.0 & 78.7 & 76.16 \\
    	\footnotesize{Colorization} & \textbf{80.0}    & \textbf{79.7} & 79.5 & 79.2 & 79.2 & 79.52 & 57.0 & 63.2 & 66.4 & 63.4 & 63.4 & 62.68 & 73.5 & 79.3 & 80.6 & 82.4 & 82.6 & 79.68 \\
    	\footnotesize{Split-Brain} & 79.5 & \textbf{79.7} & 79.5 & 80.1 & 79.4 & 79.64 & 58.2 & \underline{64.4} & \underline{67.8} & 63.8 & 65.6 & 63.96 & \textbf{77.5} & \textbf{83.7} & \underline{84.7} & \underline{83.7} & \underline{84.5} & 82.82 \\
    	\footnotesize{Counting} & 53.0 & 52.2 & 63.3 & 65.5 & 58.8 & 58.56 & \underline{61.8} & 61.3 & 60.4 & 62.8 & 62.3 & 61.72 & 75.6 & 76.0 & 74.3 & 72.5 & 71.8 & 74.04 \\
    	\footnotesize{RotNet} & 77.6 & 73.8 & \textbf{80.3} & \underline{80.3} & \underline{80.3} & 78.46 & 54.6 & 57.8 & 64.2 & \underline{66.0} & \textbf{66.0} & 61.72 & 67.2 & 66.8 & 63.6 & 63.6 & 63.6 & 64.96 \\
    	\hline
    	\footnotesize{Ours} & 78.4 & \textbf{79.9} & \underline{80.5} & \color{black}\textbf{80.8} & \textbf{80.6} & 80.02 & \textbf{62.6} & \textbf{65.9} & \textbf{68.4} & \color{black}\textbf{68.9} & \underline{65.8} & 66.32 & \underline{77.4} & \underline{83.4} & \textbf{85.3} & \textcolor{black}{\bf 85.6} & \textbf{85.1} & 83.36 \\
    	\bottomrule
    \end{tabular}
    }
    }
    \label{Tab:ResultsOnThreeBenchmarks}
\end{table*}

\section{Results and Discussions}
\label{sec:results_and_discussions}

\subsection{Evaluation of Unsupervised Features}
\label{sec:experimental_results}
\label{sec:experiments_ava}
\label{sec:AVA_Experiments}

Our experimental results on the three benchmarks are reported in Table~\ref{Tab:ResultsOnThreeBenchmarks}.
In each column, the best numbers are shown in \textbf{bold} font and the second best are highlighted with an \underline{underscore}.
We can make several interesting observations from the table.

\begin{figure}
    \centering
    \includegraphics[width=\linewidth]{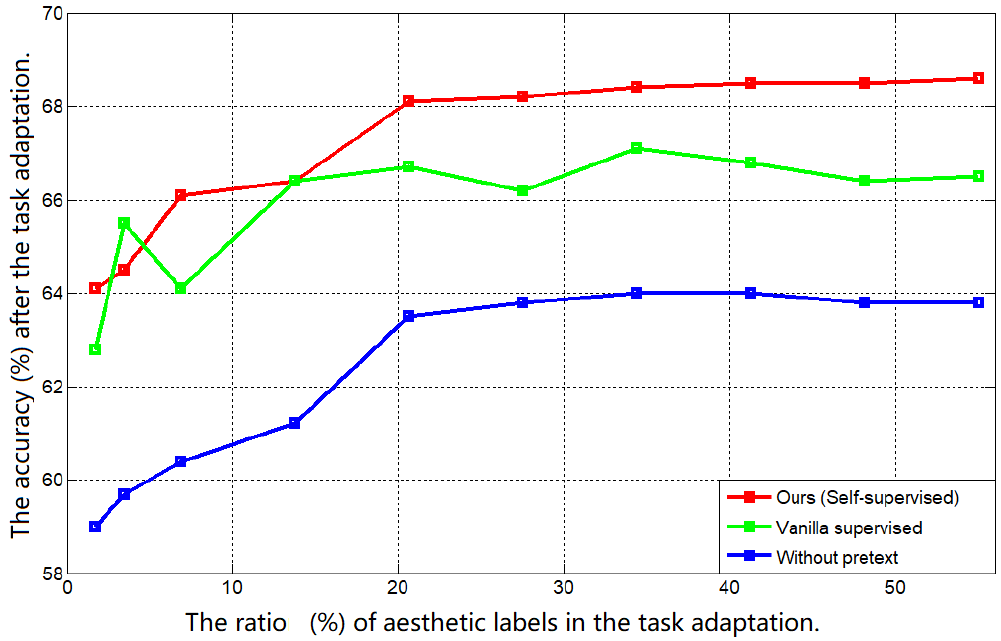}
    \vspace{-3mm}
    \caption{The accuracy results of the 3rd \emph{conv}. block from different learning schemes in low data adaptation on AADB.
    }
    \label{fig:AccuracyGoesWithNTraining_AADB}
\end{figure}

\emph{The proposed scheme generally works the best on all the three benchmarks.}
It is evident that our approach can achieve competitive results consistently, compared with other baselines on the three datasets, especially for the mid-level layers, \emph{e.g.}, $conv4$.
Self-supervised visual features can even outperform semantics-related features that are pre-trained with manual labels from ImageNet or Places.
Furthermore, from the accuracy perspective, our method can be comparable to or work better than existing self-supervised learning schemes in image aesthetic assessment.

\begin{figure}
    \centering
    \includegraphics[width=\linewidth]{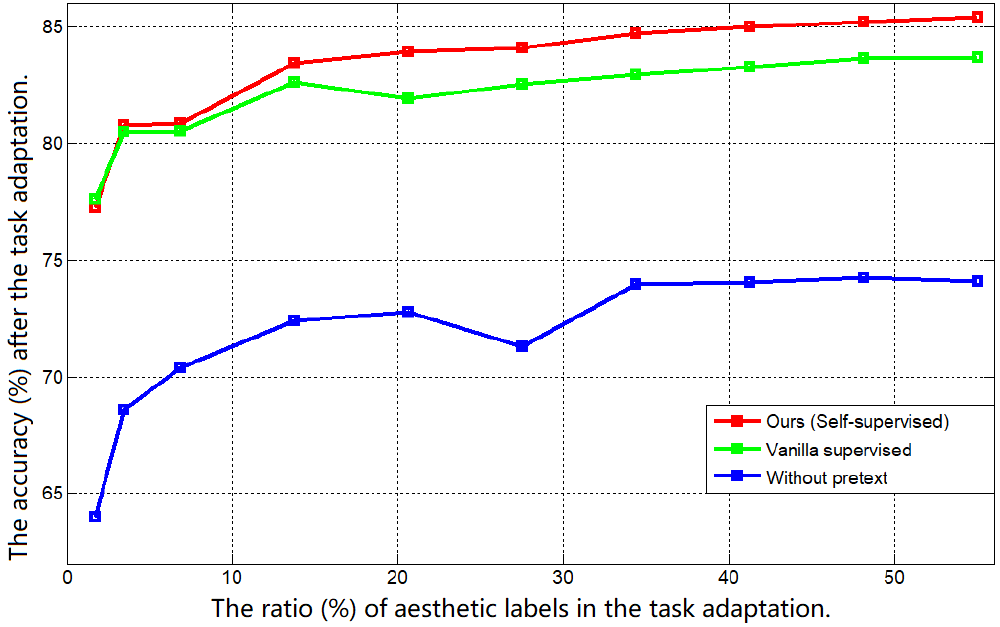}
    \vspace{-3mm}
    \caption{The accuracy results of the 3rd \emph{conv}. block from different methods in low data adaptation on CUHK-PQ.
    }
    \label{fig:AccuracyGoesWithNTraining_CUHKPQ}
\end{figure}

\emph{Mid-level features achieve the best results in task adaptation.}
By comparing performance of different layers, we can see the correlation between image aesthetics and network depth.
As shown in all tables, mid-level features ($conv3$ \& $conv4$) generally outperform high-level features ($conv5$) and low-level ones ($conv1$ \& $conv2$) in terms of accuracy.
The observations are consistent with~\cite{zhang2016colorful,gidaris2018unsupervised}.
One possible explanation is that, during the back propagation process, $conv5$ gets more training signals for the pretext task as compared to $conv4$. Consequently, $conv5$ is more likely to suffer from the overfitting issue, while using $conv4$ leads to better generalization performance.

\emph{In addition to image attributes used in our pretext tasks, color is another important factor in image aesthetic assessment.}
Among other tested pretext tasks, Split-brain~\cite{zhang2017split} and Colorization~\cite{zhang2016colorful} consistently achieve the better results.
It indicates that color and image attributes in Tab.~\ref{Tab:DegradationType} are the key factors in assessing visual aesthetics.
Regarding why RotNet fails to achieve good results on the CUHK-PQ benchmark, we suspect that both high-quality images and low-quality ones in this dataset share similar distributions or visual patterns\commentout{ in the structure line}.

\begin{table*}
    \centering
    \footnotesize{
    \caption{Results of several methods on AVA benchmark measured in terms of binary classification accuracy of aesthetic labels.}
    \label{Tab:ResultsOnAVA_VSFS}
    \begin{tabular}{r|cccc}
    	\toprule
    	\multirow{2}{*}{Method} & \multirow{2}{*}{Backbone} & Labels in & Aesthetic & \multirow{2}{*}{Results ($\%$)} \\
    	 & & the pre-training & label alone & \\
    	\hline
    	DMA-Net~\cite{lu2015deep}            & AlexNet & \multirow{8}{*}{$\sim 10M$} & $\times$ & 75.4 \\
    	AA-Net~\cite{wang2017deep}               & VGG & & $\times$ & 76.9 \\
    	~\cite{kong2016photo}                    & AlexNet & & $\times$ & 77.3 \\
    	MNA-CNN~\cite{mai2016composition}  & VGG & & $\times$ & 77.4 \\
    	NIMA~\cite{talebi2018nima}               & Inception & & $\surd$ & 81.5 \\
    	Pool-3FC~\cite{hosu2019effective}        & Inception & & $\surd$ & 81.7 \\
    	A-Lamp~\cite{ma2017lamp}                 & VGG & & $\times$ & 82.5 \\
    	$MP_{ada}$~\cite{sheng2018attention}     & ResNet-18 & & $\surd$ & 83.0 \\
    	\hline
    	Our pretext task & AlexNet & \multirow{2}{*}{0} & \multirow{2}{*}{$\surd$} & 82.0 \\
    	+ non-linear layers & ResNet-18 & & & 82.8 \ \\
    	\bottomrule
    \end{tabular}
    }
\end{table*}

\subsection{Low Data Adaptation}
One practical issue that self-supervised learning schemes are able to handle is low data adaptation, where very few labels are available for task adaptation.
In our case, we simulate the low data adaptation regime by using $5\% \sim 50\%$ of the original training data per class in the task adaptation stage.
The final results are shown in Fig.~\ref{fig:AccuracyGoesWithNTraining_AADB} and Fig.~\ref{fig:AccuracyGoesWithNTraining_CUHKPQ}.
From these two figures, we can see that our proposed learning scheme generally outperforms baseline models pre-trained with $1,000$-ways labels from ImageNet.
As the margin becomes larger when using more manual aesthetic labels in the task adaptation, our method presents higher efficiency of data usage compared to the vanilla supervised counterpart.

It is also interesting to note that our method and vanilla supervised method have similar adaptation performance when $1\% \sim 10 \%$ training data is available.
We believe this fact is due to the complexity of visual aesthetics, i.e., when aesthetic labels are extremely few, the sampled instances cannot cover the entire distribution faithfully and thus lead to poor assessment results.

\subsection{Adaption Using a Non-Linear Classifier}
If we use non-linear layers in the task adaptation stage, we can achieve results close to state-of-the-art fully supervised approaches~\cite{ma2017lamp,sheng2018attention,talebi2018nima,hosu2019effective} which have a test accuracy of $80\%\sim83\%$ on the AVA benchmark, as shown in Tab.~\ref{Tab:ResultsOnAVA_VSFS}.
Note that our method does not use $10$ millions labels from ImageNet during the pre-training, and we only use aesthetic labels in the task adaptation stage.
Besides, our accuracy is about $69.2\%$ on the AADB benchmark and $88.5\%$ on the CUHKPQ dataset, using ResNet-18 as the backbone network.
These numbers are also close to that of the same network pre-trained with $10$ millions labels from ImageNet (\emph{i.e.}, i.e., $70.4\%$ on the AADB and $90.2\%$ on the CUHKPQ).
Arguably, we manage to achieve similar assessment performance using far less manual labels.

\subsection{Ablation Study}
\label{sec:ablation_study}

\paragraph{Pretext tasks.}
We perform the pre-training with either one of the two pretext tasks and measure the final assessment results.
It turns out that joint training using both loss terms generally yields better aesthetic assessment accuracy than using $\degradationloss(\patch, \theta_{\transformation})$ or $\tripletloss(\patch, \theta_{\transformation_1}, \theta_{\transformation_2})$ alone.
The accuracy difference is $\sim 0.3\%$ on the AVA dataset and $\sim 0.7\%$ on the AADB dataset,.
We also find that pre-training with $\tripletloss(\patch, \theta_{\transformation_1}, \theta_{\transformation_2})$ alone is prone to undesirable training dynamics, while the two-stage learning scheme is more stable and consequently leads to better results.

\paragraph{Image editing operations.}
We randomly select one operation in Tab~\ref{Tab:DegradationType} and exclude it from the pre-training stage to analyze the impact on the final assessment performance.
We can make several observations from the corresponding results shown in Tab.~\ref{tab:AblationStudyOfOperations}.
First, editing operations which are related to softness, camera shake, and poor lighting are relatively more significant in learning aesthetic-aware features, compared to other operations.
Second, image manipulations which are related to distracting, fuzziness, and noise have relatively smaller margins.
This fact indicates that these operations may create aesthetically ambiguous instances that do not always provide consistent signals.
Interestingly, camera shake has the most significant impact on the AVA, while soft/grainy are the most important ones for the AADB.

\begin{table}
    \centering
    \caption{The degradation caused by discarding some image manipulation operations during the pretext task training.}
    \label{tab:AblationStudyOfOperations}
        \begin{tabular}{l|c|c}
        \toprule
        Attribute & AVA & AADB \\
        \hline
        Much noise         & $\thicksim 0.1\%$ & $\thicksim 0.3\%$ \\
        Poor lighting       & $\thicksim 0.3\%$ & $\thicksim 0.4\%$ \\
        Soft / Grainy       & $\thicksim 0.4\%$ & $0.5 \thicksim 0.8\%$ \\
        Fuzzy                  & $\thicksim 0.1\%$ & $\thicksim 0.2\%$ \\
        Camera shake    & $0.3 \thicksim 0.5\%$ & $0.4 \thicksim 0.6\%$ \\
        Distracting           & $\thicksim 0.2\%$ & $0.3 \thicksim 0.4\%$ \\
        \bottomrule
    \end{tabular}
\end{table}

\paragraph{Entropy-based weighting.}
Without our entropy-based weighting scheme, there will be at least $0.5\%$ down in the performance after task adaptation.
Besides, some undesired dynamics will occur during the training process.
Therefore, we can strengthen meaningful training signals from manipulated instances and improve training efficiency by assigning an entropy-based weight to each patch.

\section{Conclusions}
In this paper, we propose a novel self-supervised learning scheme to investigate the possibility of learning useful aesthetic-aware features without manual annotations.
Based on the correlation between negative aesthetic effects and several expert-designed image manipulations, we argue that an aesthetic-aware representation space should distinguish between results yielded by various operations.
To this end, we propose two pretext tasks, one to recognize what kind of editing operation has been applied to an image patch, and the other to capture fine-grained aesthetic variations due to different manipulation parameters.
Our experimental results on three benchmarks demonstrate that the proposed scheme can learn aesthetics-aware features effectively and generally outperforms existing self-supervised counterparts.
Besides, we achieve results comparable to state-of-the-art supervised methods on the AVA dataset without using labels from ImageNet.
We hope these findings can help us obtain a better understanding of image visual aesthetics and inspire future research in related areas.

\section*{Acknowledgements}
\footnotesize{This work was supported in part by National Natural Science Foundation of China under nos. 61832016, 61672520 and 61720106006, in part by Key R\&D Program of Jiangxi Province (No. 20171ACH80022), in part by CASIA-Tencent Youtu joint research project.}

\bibliography{SSAVA_AAAI}
\bibliographystyle{aaai}

\end{document}